\documentclass[a4paper,oneside,12pt,english]{article}
\usepackage{url}
\usepackage[pdftex]{color,graphicx}
\usepackage[backref=false]{hyperref}
\hypersetup{plainpages = false,
              breaklinks=true,
              linktocpage,
              colorlinks   = true, 
              urlcolor     = blue, 
              linkcolor    = blue, 
              citecolor   = red, 
              pdfauthor={Niko Brummer},
              pdfstartpage=1, 
              pdfstartview=FitH,  
              pdfkeywords={}}

\usepackage{amsmath}%
\usepackage{amsfonts}%
\usepackage{amssymb}%
\usepackage{graphicx}

\usepackage{tikz}
\usetikzlibrary{bayesnet}

\usepackage{enumerate}

\def\lambdavec{\boldsymbol{\lambda}}

\def\zvec{\mathbf{z}}

\def\ND{\mathcal{N}}

\def\R{\mathbb{R}}

\begin{document}
\title{Note on the equivalence of hierarchical variational models and auxiliary deep generative models}
\author{Niko Br\"ummer\\ AGNITIO Research South Africa}
\date{Strand, March 2016}
\maketitle

\begin{abstract}
This note compares two recently published machine learning methods for constructing flexible, but tractable families of variational hidden-variable posteriors. The first method, called \emph{hierarchical variational models} enriches the inference model with an extra variable, while the other, called \emph{auxiliary deep generative models}, enriches the generative model instead. We conclude that the two methods are mathematically equivalent.   
\end{abstract}

\section{Introduction}
\def\xvec{\mathbf{x}}
\def\VBL{\mathcal{L}}
In machine learning, there is an ongoing revival of the use of variational Bayes (VB) to deal with complex probabilistic models with hidden variables. The revival is driven by the use of stochastic methods to approximate the VB lower bound and associated gradients. See for example~\cite{AEVB,Google,Titsias}. The advantages include automated inference~\cite{ADVI} and also that they are applicable to a much wider class of probabilistic models. While the basic recipes are limited in the flexibility of the approximate hidden-variable posteriors, there are ongoing efforts to make them more flexible. For example, \cite{flows} allows complex reparametrizations of the hidden variables. These reparametrizations are however still limited by an invertibility constraint and by requiring computation of Jacobian determinants. This note summarizes and compares further progress in this regard as published in~\cite{HVM,VGP,ADGM}.  

\section{Problem statement}
We start by defining of the kind of probabilistic models of interest, introduce the stochastic VB solution to deal with them and state a limitation of stochastic VB w.r.t.\ the approximate posteriors. 

Our model has the following form. Let $\xvec$ be observed and $\zvec$ be hidden. The generative model is
\begin{align}
P(\xvec,\zvec) &= P(\xvec\mid\zvec)P(\zvec)
\end{align}
where we assume we can evaluate both factors in the RHS and therefore also the LHS. However, the marginal $P(\xvec)$ and posterior $P(\zvec|\xvec)$ are assumed intractable. The model, $P(\xvec,\zvec)$, will usually be conditioned on some parameters, but these parameters play no role in this discussion and are considered implicit in the definition of $P$. 

Variational Bayes (VB) allows us to employ some convenient family of parametrized distributions, $Q(\zvec|\theta)$, where $\theta$ can be chosen to approximate the intractable posterior:
\begin{align}
Q(\zvec\mid\theta) &\approx P(\zvec\mid\xvec)
\end{align}
In the standard VB recipe, $\theta$ is found by maximizing the evidence lower bound (ELBO):
\begin{align}
\label{eq:ELBO}
\VBL(\theta) &= \int Q(\zvec\mid\theta) \log\frac{P(\xvec,\zvec)}{Q(\zvec\mid\theta)} \,d\zvec
\end{align}
If the ELBO integral cannot be evaluated in closed form, it can still be approximated stochastically by using, say $N$, samples from $Q(\zvec|\theta)$:
\begin{align}
\VBL(\theta) &\approx \frac1N \sum_{\zvec\sim Q(\zvec|\theta)} \log\frac{P(\xvec,\zvec)}{Q(\zvec\mid\theta)} 
\end{align}
The gradients w.r.t.\ $\theta$, which are needed for the optimization, require some further tricks, for example the reparametrization trick~\cite{AEVB}. See~\cite{Titsias} for a review of methods to deal with the gradients.   

The above VB recipe can only be applied if we can evaluate $Q(\zvec|\theta)$. Below we summarize two published methods, \emph{hierarchical variational models}~\cite{HVM,VGP} and \emph{auxiliary deep generative models}~\cite{ADGM} that construct approximate posteriors that cannot be evaluated, but which can nevertheless be used in slightly more complex VB strategies.

\section{Hierarchical Variational Models}
This is a summary and discussion of ideas used in~\cite{HVM,VGP}. The VB posterior is assembled by involving another hidden variable, say $\lambdavec$, which must be marginalized out because it plays no role in the generative model $P(\xvec,\zvec)$. The full inference model is defined as:
\begin{align}
Q(\zvec,\lambdavec\mid\theta) &= Q(\lambdavec\mid\theta) Q(\zvec\mid\lambdavec,\theta) 
\end{align}
We assume that we can evaluate and sample from both factors in the RHS. This means the joint density (LHS) can also be evaluated and sampled from---i.e.\ by doing ancestral sampling. The marginal
\begin{align}
Q(\zvec\mid\theta) &= \int Q(\zvec,\lambdavec\mid\theta) \,d\lambdavec
\end{align}
is however assumed to be intractable---and by implication also the posterior $Q(\lambdavec|\zvec,\theta)$. 

We deal with this inference model via a nested VB recipe, where we introduce yet another approximate posterior:
\begin{align}
R(\lambdavec\mid\zvec,\phi) &\approx Q(\lambdavec\mid\zvec,\theta)
\end{align} 
that is chosen from a family of distributions that we can evaluate.\footnote{We don't need to be able to sample from it.} We now modify the ELBO by subtracting a non-negative term, namely an expected KL divergence, so that the \emph{modified ELBO} is still a lower bound to the evidence:
\begin{align}
\label{eq:ELBO2}
\VBL(\theta,\phi) &= \VBL(\theta) - \int Q(\zvec\mid\theta) \left[
\int Q(\lambdavec\mid\zvec,\theta)\log\frac{Q(\lambdavec\mid\zvec,\theta)}{R(\lambdavec\mid\zvec,\phi)}\,d\lambdavec
\right] \,d\zvec 
\end{align}
where the factor in square brackets is the KL-divergence. If we maximize $\VBL(\theta,\phi)$ w.r.t.\ $\phi$, it will be forced closer to the original $\VBL(\theta)$, reaching equality if $R(\lambdavec|\zvec,\phi) = Q(\lambdavec|\zvec,\theta)$ for every value of $z$ in the support of $Q(\zvec|\theta)$. By this construction we now have:
\begin{align}
\label{eq:elbos}
\log P(\xvec) \ge \VBL(\theta) \ge \VBL(\theta,\phi).
\end{align}
We now re-arrange $\VBL(\theta,\phi)$ in terms of tractable components. Since we can sample from $Q(\zvec,\lambdavec|\theta)$, we are happy with expectations formed over $\zvec,\lambdavec$. So we can rewrite:
\begin{align}
\label{eq:ELBO3}
\begin{split}
\VBL(\theta,\phi) 
&= \VBL(\theta) + \int\int Q(\zvec,\lambdavec\mid\theta)\log\frac{R(\lambdavec\mid\zvec,\phi)}{Q(\lambdavec\mid\zvec,\theta)}\,d\lambdavec \,d\zvec 
\end{split} 
\end{align}
Since the integrand in~\eqref{eq:ELBO} is not a function of $\lambda$, it can be rewritten as:
\begin{align}
\label{eq:ELBO4}
\VBL(\theta) &= \int\int Q(\zvec,\lambdavec\mid\theta) \log\frac{P(\xvec,\zvec)}{Q(\zvec\mid\theta)} \,d\lambdavec \,d\zvec
\end{align}
Both~\eqref{eq:ELBO3} and~\eqref{eq:ELBO4} have intractable factors, namely $Q(\lambdavec|\zvec,\theta)$ and $Q(\zvec|\theta)$. But when we combine them, these factors are multiplied to form the joint distribution, $Q(\zvec,\lambdavec\mid\theta)$, which we \emph{can} evaluate. This gives:
\begin{align}
\label{eq:ELBO5}
\begin{split}
\VBL(\theta,\phi) &= \int\int Q(\zvec,\lambdavec\mid\theta) \left[
\log\frac{P(\xvec,\zvec)}{Q(\zvec\mid\theta)} 
+\log\frac{R(\lambdavec\mid\zvec,\phi)}{Q(\lambdavec\mid\zvec,\theta)}\right]\,d\lambdavec \,d\zvec \\
&=\int\int Q(\zvec,\lambdavec\mid\theta) 
\log\frac{P(\xvec,\zvec)R(\lambdavec\mid\zvec,\phi)} 
{Q(\zvec,\lambdavec\mid\theta)}\,d\lambdavec \,d\zvec  \\
&\approx \frac{1}{N}\sum_{\zvec,\lambdavec\sim Q(\zvec,\lambdavec\mid\theta)} \log\frac{P(\xvec,\zvec)R(\lambdavec\mid\zvec,\phi)} 
{Q(\zvec,\lambdavec\mid\theta)}
\end{split} 
\end{align}

\subsection{Example}
\def\nulvec{\boldsymbol{0}}
\def\Imat{\mathbf{I}}
We can use this recipe to involve a general-purpose feed-forward neural net to define a flexible family for the variational posterior. Let $\zvec=(z_1,\ldots,z_n)\in\R^n$ and let $\lambdavec\in\R^m$. Let: 
\begin{align}
Q(\zvec\mid\lambdavec,\theta) &= \prod_{i=1}^n\ND\bigl(z_i\mid\mu_i(\lambdavec;\theta),\sigma_i^2(\lambdavec;\theta)\bigr)
\end{align}
where we employ a neural net, parametrized by $\theta$, to map $\lambdavec$ to the means and variances of a diagonal Gaussian distribution for $\zvec$. Let
\begin{align}
Q(\lambdavec\mid\theta)&=Q(\lambdavec)=\ND(\lambdavec\mid\nulvec,\Imat)
\end{align}
This gives a very flexible continuous \emph{Gaussian mixture}:
\begin{align}
Q(\zvec\mid\theta) &= \int \ND(\lambdavec\mid\nulvec,\Imat)\prod_{i=1}^n\ND\bigl(z_i\mid\mu_i(\lambdavec;\theta),\sigma_i^2(\lambdavec;\theta)\bigr) \,d\lambdavec
\end{align}
We can sample from it using ancestral sampling, but we cannot evaluate the marginal density $Q(\zvec\mid\theta)$. This makes the above recipe applicable.

We can form $R(\lambdavec|\zvec,\phi)$ similarly with a neural net, parametrized by $\phi$, that maps $\zvec$ to the parameters of a Gaussian distribution for $\lambdavec$.

\section{Auxiliary Deep Generative Models}
We now summarize the other method, published in~\cite{ADGM} and show that it is very close to the method in~\cite{HVM,VGP}. Let's recall~\eqref{eq:ELBO5}:
\begin{align}
\VBL(\theta,\phi) 
&=\int\int Q(\zvec,\lambdavec\mid\theta) 
\log\frac{P(\xvec,\zvec)R(\lambdavec\mid\zvec,\phi)} 
{Q(\zvec,\lambdavec\mid\theta)}\,d\lambdavec \,d\zvec  
\end{align}
where $P$ belongs to the generative model and $Q,R$ to the inference model. However, nothing changes mathematically if we rename $R$ to $P$ and thereby change the role of $\lambda$ to be a hidden variable in both generative and inference models:
\begin{align}
\label{eq:ELBO6}
\begin{split}
\VBL(\theta,\phi) 
&=\int\int Q(\zvec,\lambdavec\mid\theta) 
\log\frac{P(\xvec,\zvec)P(\lambdavec\mid\zvec,\phi)} 
{Q(\zvec,\lambdavec\mid\theta)}\,d\lambdavec \,d\zvec  \\
&=\int\int Q(\zvec,\lambdavec\mid\theta) 
\log\frac{P(\xvec,\zvec,\lambdavec\mid\phi)} 
{Q(\zvec,\lambdavec\mid\theta)}\,d\lambdavec \,d\zvec  
\end{split}
\end{align}
which is just the standard ELBO for an \emph{extended model}, $P(\xvec,\zvec,\lambdavec|\phi)$, with two hidden variables, $\zvec$ and $\lambdavec$. If the original model, $P(\xvec,\zvec)=P(\xvec|\zvec)P(\zvec)$ is visualized as:
$$
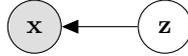
\begin{tikzpicture}
\node[obs]  (x) {$\xvec$};
\node[latent, right =of x] (z) {$\zvec$};
\edge {z}{x};
\end{tikzpicture}$$
Then the extended model implied by~\eqref{eq:ELBO6}, is 
$$
P(\xvec,\zvec,\lambdavec|\phi)=P(\xvec\mid\zvec)P(\zvec)P(\lambdavec\mid\zvec,\phi)
$$ 
and can be visualized as:
$$
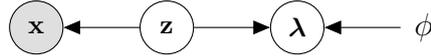
\begin{tikzpicture}
\node[obs]  (x) {$\xvec$};
\node[latent, right =of x] (z) {$\zvec$};
\node[latent, right =of z] (lambda) {$\lambdavec$};
\node[const, inner sep = 5pt, right =of lambda] (phi) {$\phi$};
\edge {z}{x};
\edge {z}{lambda};
\edge {phi}{lambda};
\end{tikzpicture}$$
Since $\lambdavec$ is hidden, the original variables $\zvec$ and $\xvec$ are independent of $\phi$, so that $\phi$ may be modified at will, without changing the generative description of the data: $P(\xvec|\phi)=P(\xvec)$. As the analysis in the previous section showed, by maximizing~\eqref{eq:ELBO6} w.r.t.\ $\phi$, we are instead effectively improving the \emph{inference} model.

As pointed out in both~\cite{VGP} and ~\cite{ADGM}, it turns out this recipe still works for a slightly more complex extended model, visualized as:
$$
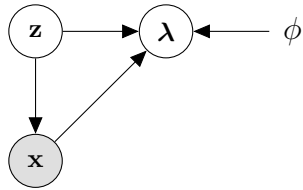
\begin{tikzpicture}
\node[obs]  (x) {$\xvec$};
\node[latent, above =of x] (z) {$\zvec$};
\node[latent, right =of z] (lambda) {$\lambdavec$};
\node[const, inner sep = 5pt, right =of lambda] (phi) {$\phi$};
\edge {z}{x};
\edge {z}{lambda};
\edge {phi}{lambda};
\edge {x}{lambda};
\end{tikzpicture}$$
That is, we have replaced $R(\lambdavec|\zvec,\phi)$ by $R(\lambdavec|\zvec,\xvec,\phi)$. It should be clear from~\eqref{eq:ELBO2} that this can be done without changing the relationship~\eqref{eq:elbos} between the standard and modified lower bounds. 

\section{Conclusion}
Mathematically, the two methods are the same. Perhaps the derivation of the method is easier to understand from the \emph{hierarchical variational model} viewpoint, while software implementation could be facilitated by the \emph{auxiliary deep generative models} viewpoint.

\bibliographystyle{IEEEtran}

\begin{thebibliography}{10}
\bibitem[1] {AEVB} Diederik P Kingma and Max Welling``Auto-Encoding Variational Bayes'', December 2013. \url{http://arxiv.org/abs/1312.6114}.
\bibitem[2] {Google} Danilo Jimenez Rezende, Shakir Mohamed and Daan Wierstra, ``Stochastic backpropagation and approximate
inference in deep generative models'', January 2014. \url{http://arXiv:1401.4082, 2014}.
\bibitem[3] {ADVI} Alp Kucukelbir, Dustin Tran, Rajesh Ranganath, Andrew Gelman and David M. Blei, ``Automatic Differentiation Variational Inference'', March 2016. \url{http://arxiv.org/abs/1603.00788}.
\bibitem[4] {flows} Danilo Jimenez Rezende, Shakir Mohamed, ``Variational Inference with Normalizing Flows'', May 2015. \url{http://arxiv.org/abs/1505.05770}.
\bibitem[5] {Titsias} Michalis K. Titsias, ``Local Expectation Gradients for Doubly Stochastic Variational Inference'', March 2015/ \url{http://arxiv.org/abs/1503.01494}.
\bibitem[6] {HVM} Rajesh Ranganath, Dustin Tran and David M. Blei, ``Hierarchical Variational Models'', November 2015. \url{http://arxiv.org/abs/1511.02386}.
\bibitem[7] {VGP} Dustin Tran, Rajesh Ranganath and David M. Blei, ``Variational Gaussian Process'', November 2015. \url{http://arxiv.org/abs/1511.06499}.
\bibitem[8] {ADGM} Lars Maal{\o}e, Casper Kaae S{\o}nderby, S{\o}ren Kaae S{\o}nderby, Ole Winther, ``Auxiliary Deep Generative Models'', February 2016. \url{http://arxiv.org/abs/1602.05473}.
\end{thebibliography}

\end{document}